\documentclass{article}
\PassOptionsToPackage{numbers, compress}{natbib}

     \usepackage[preprint]{neurips_2020}

\usepackage[utf8]{inputenc} % allow utf-8 input
\usepackage[T1]{fontenc}    % use 8-bit T1 fonts
\usepackage{hyperref}       % hyperlinks
\usepackage{url}            % simple URL typesetting
\usepackage{booktabs}       % professional-quality tables
\usepackage{amsfonts}       % blackboard math symbols
\usepackage{nicefrac}       % compact symbols for 1/2, etc.
\usepackage{microtype}      % microtypography
\usepackage{bm}
\usepackage{natbib}

\usepackage{rotating}

\usepackage{makecell}
\usepackage{tabularx}
\usepackage{caption, subcaption, floatrow, wrapfig}
\newfloatcommand{capbtabbox}{table}[][\FBwidth]

\usepackage{amsmath}
\usepackage{graphicx}

\DeclareMathOperator*{\argmin}{arg\,min}
\newcommand\norm[1]{\left\lVert#1\right\rVert}

\title{Score-Guided Generative Adversarial Networks}

\author{%
  Minhyeok Lee\\
  College of Engineering\\
  Korea University\\
  Seoul, Korea 02841\\
  \texttt{suam6409@korea.ac.kr}\\
  \And
  Junhee Seok\thanks{Corresponding author.}\\
  School of Electrical Engineering\\
  Korea University\\
  Seoul, Korea 02841\\
  \texttt{jseok14@korea.ac.kr}\\
}

\begin{document}

\maketitle

\begin{abstract}
We propose a Generative Adversarial Network (GAN) that introduces an evaluator module using pre-trained networks. The proposed model, called score-guided GAN (ScoreGAN), is trained with an evaluation metric for GANs, i.e., the Inception score, as a rough guide for the training of the generator. By using another pre-trained network instead of the Inception network, ScoreGAN circumvents the overfitting of the Inception network in order that generated samples do not correspond to adversarial examples of the Inception network. Also, to prevent the overfitting, the evaluation metrics are employed only as an auxiliary role, while the conventional target of GANs is mainly used. Evaluated with the CIFAR-10 dataset, ScoreGAN demonstrated an Inception score of 10.36$\pm$0.15, which corresponds to state-of-the-art performance. Furthermore, to generalize the effectiveness of ScoreGAN, the model was further evaluated with another dataset, i.e., the CIFAR-100; as a result, ScoreGAN outperformed the other existing methods, where the Fréchet Inception Distance (FID) was 13.98.
\end{abstract}

\section{Introduction}

Due to its innovative training algorithm and superb performance in image generation tasks, Generative Adversarial Networks (GANs) have been widely studied in recent years \cite{RN1,RN2,RN3,RN32}. GANs generally employ two Artificial Neural Network (ANN) modules, called a generator and a discriminator, which are trained with an adversarial process to detect and deceive each other. Specifically, the discriminator aims at detecting synthetic samples that are produced by the generator; meanwhile, the generator is trained by errors that are obtained from the discriminator. By such a competitive learning process, the generator can produce fine synthetic samples of which features are incredibly similar to those of actual samples \cite{RN4,RN5}.

However, the performance evaluation of GAN models is a challenging task since the quality and diversity of generated samples should be assessed by the human perspective \cite{RN6, RN7}; furthermore, unbiased evaluations are also difficult because each person can have different views on the quality and diversity of samples. Therefore, several studies have introduced quantitative metrics to evaluate GAN models in a measurable manner \cite{RN7, RN8}.

The Inception score is one of the most representative metrics to evaluate GAN models for image generation \cite{RN7}. A conventional pre-trained ANN model for image classification, called Inception network \cite{RN9}, is employed to assess both quality and diversity of generated samples, by measuring entropies of inter- and intra-samples in terms of estimated probabilities for each class. Fréchet Inception Distance (FID) is another metric to measure GAN performance, in which the distance between feature distributions of real samples and generated samples is calculated \cite{RN8}.

From the adoption of the evaluation metrics, the following questions then arise: \emph{Can the evaluation metrics be used as targets for the training of GAN models} since the metrics reasonably represent the quality and diversity of samples? By backpropagating gradients of the score or distance, is it possible to maximize or minimize them? Such an approach seems feasible since the metrics are generally differentiable; therefore, the gradients can be computed and backpropagated.

However, simply backpropagating the gradients and training with the metrics correspond to learning adversarial examples in general \cite{RN10, RN11}. Since the complexity of ANN models is significantly high, we can easily make a sample to be incorrectly predicted, by adding minimal noises into the sample; this noisy sample is called the adversarial example \cite{RN11}. Therefore, in short, fine quality and rich diversity of samples can have a high Inception score, while the reverse is not always true.

Barratt and Sharma \cite{RN12} have studied this problem and found that directly maximizing the score does not guarantee that the generator produces fine samples. In experiments in the study, a GAN model is trained to maximize the Inception score; then, the trained model produces image samples with a very high Inception score. While the Inception score of real samples in the CIFAR-10 dataset is around 10.0, the produced images achieve an Inception score of 900.15 \cite{RN12}. However, the produced images are entirely different from real images in the CIFAR-10 dataset; instead, they look like noises.

In this paper, to address such a problem and utilize the evaluation metric as a training method, we propose a score-guided GAN (ScoreGAN) that employs an evaluator ANN module using pre-trained networks with the evaluation metrics. While the aforementioned problems exist in ordinary GAN, ScoreGAN solves the problems through two approaches as follows.

First, ScoreGAN uses the evaluation metric as an auxiliary target, while the target function of ordinary GANs is mainly used. Using the evaluation metric as the only target causes overfitting the network used for the metric, instead of learning meaningful information from the network, as shown in related studies \cite{RN12}. Thus, the evaluation metric is employed as the auxiliary target in ScoreGAN.

Second, for backpropagating gradients and training the generator in ScoreGAN, we employ the other pre-trained model, called MobileNet \cite{RN13}, instead of the Inception network, which can make the generator not to overfit the Inception network; also, with such an approach, we can validate that the generator actually learns features and does not overfit the Inception network by verifying that \emph{ScoreGAN achieves a high Inception score without using the Inception network as a target.}

The main contributions of this paper are as follows:
\begin{itemize}
  \item The score-guided GAN (ScoreGAN) that uses the evaluation metric as an additional target of GAN is proposed
  \item The proposed ScoreGAN circumvents the overfitting problem as well as making adversarial examples by not using the Inception network as the evaluator
  \item Evaluated by the Inception score and cross-validated through the FID, ScoreGAN demonstrates state-of-the-art performance over the CIFAR-10 dataset and CIFAR-100 dataset, in which its Inception score in the CIFAR-10 is 10.36$\pm$0.15, and the FID in the CIFAR-100 is 13.98.
\end{itemize}

\section{Background}
\subsection{Controllable generative adversarial networks}

An ordinary GAN model consists of two ANN modules, i.e., the generator and the discriminator. The two modules are trained by playing a game to deceive or detect each other \cite{RN6,RN14}. The game to train a GAN can be represented as follows:
\begin{equation}
\hat{\bm{\theta}}_{D} = \argmin_{\bm{\theta}_{D}}\left\{L_{D}\left(1,D\left(X;\bm{\theta}_{D}\right)\right)+L_{D}\left(0,D\left(G\left(Z;\hat{\bm{\theta}}_{G}\right);\bm{\theta}_{D}\right)\right)\right\},
\end{equation}
\begin{equation}
\hat{\bm{\theta}}_{G} = \argmin_{\bm{\theta}_{G}}\left\{L_{D}\left(1,D\left(G\left(Z;\bm{\theta}_{G}\right);\bm{\theta}_{D}\right)\right)\right\},
\end{equation}
where $G$ and $D$ denote the generator and the discriminator, respectively, $X$ is a training sample, $Z$ represents a noise vector, $\bm{\theta}$ is a set of weights of an ANN model, and $L_{D}$ indicates a loss function for the discriminator.

However, the ordinary GAN can hardly produce desired samples since each feature in a dataset is randomly mapped into each variable of the input noise vector. Therefore, it is hard to discover which noise variable corresponds to which feature. To overcome this problem, conditional variants of GAN that introduce conditional input variables have been studied \cite{RN15, RN16, RN17}.

Controllable GAN (ControlGAN) \cite{RN18} is one of the conditional variants of GAN that uses an independent classifier and the data augmentation techniques to train the classifier. While a conventional model, called Auxiliary Classifier GAN (ACGAN) \cite{RN17}, has an overfitting issue on the classification loss and a trade-off for using the data augmentation technique \cite{RN18}, ControlGAN breaks the trade-off through introducing the independent classifier as well as the data augmentation technique. The training of ControlGAN is performed as follows:
\begin{equation}
\hat{\bm{\theta}}_{D} = \argmin_{\bm{\theta}_{D}}\left\{L_{D}\left(1,D\left(X;\bm{\theta}_{D}\right)\right)+L_{D}\left(0,D\left(G\left(Z,\mathcal{L};\hat{\bm{\theta}}_{G}\right);\bm{\theta}_{D}\right)\right)\right\},
\end{equation}
\begin{equation}
\label{eq:4}
\hat{\bm{\theta}}_{G} = \argmin_{\bm{\theta}_{G}}\left\{L_{D}\left(1,D\left(G\left(Z;\bm{\theta}_{G}\right);\bm{\theta}_{D}\right)\right)+\gamma_{t}\cdot L_{C}\left(\mathcal{L},C\left(G\left(Z,\mathcal{L};\bm{\theta}_{G}\right);\hat{\bm{\theta}}_{C}\right)\right)\right\},
\end{equation}
\begin{equation}
\hat{\bm{\theta}}_{C} = \argmin_{\bm{\theta}_{C}}\left\{L_{C}\left(\mathcal{L},C\left(X;\bm{\theta}_{C}\right)\right)\right\},
\end{equation}
where $C$ represents the independent classifier, $\mathcal{L}$ denotes input labels, and $\gamma_{t}$ is a learning parameter that modulates the training of the generator in terms of the classification loss.

\subsection{The Inception score}

To assess the quality and diversity of generated samples by GANs, the Inception score \cite{RN7} is one of the most conventional evaluation metrics that has been extensively employed in many studies \cite{RN1,RN5,RN7,RN12,RN15,RN16,RN18}. For the quantitative evaluation of GANs, the Inception score introduces the Inception network that is initially used for image classification \cite{RN9}. The Inception network is pre-trained to solve the image classification task over the ImageNet dataset \cite{RN19}, which contains more than one million images of 1,000 different classes; then, the network learns general features of various objects.

Through the pre-trained Inception network, the quality and diversity of generated samples can be obtained with these two aspects \cite{RN7,RN12}: First, the high quality of an image can be guaranteed if the image is firmly classified into a specific class; Second, a high entropy in the marginal probability of generated samples indicates a rich diversity of the samples since such a condition signifies that the generated samples are different each other.

Therefore, entropies of intra- and inter-sample are calculated over generated samples; then, these two entropies compose the Inception score as follows:
\begin{equation}
\label{eq:6}
    IS\left(G\left(\cdot;\bm{\hat{\theta}}_{G}\right)\right)=\exp\left(\frac{1}{N}\sum KL\left(Pr\left(Y|\hat{X}\right) || Pr\left(Y\right)\right)\right),
\end{equation}
where $\hat{X}$ denotes a generated sample, $KL$ indicates the Kullback–Leibler (KL) divergence, namely, the relative entropy, and $N$ is the number of samples in a batch. Since a high KL divergence signifies a significant difference between the two probabilities, thus, a higher Inception score indicates greater qualities and a wider variety of samples. Generally, ten sets, each of which contains 5,000 generated samples, are used to calculate the Inception score \cite{RN7,RN12}.

\subsection{The Fréchet Inception distance (FID)}
The FID is another metric to evaluate generated samples in which the Inception network is employed as well \cite{RN8}. Instead of the predicted probabilities, the FID introduces the feature distribution of generated samples that can be represented as outputs of the penultimate layer of the Inception network.

With the assumption that the feature distribution follows a multivariate normal distribution, the distance between the feature distributions of real samples and generated samples is calculated as follows:
\begin{equation}
    FID\left(\bm{X},\hat{\bm{X}}\right)=\norm{\mu_{\bm{X}}-\mu_{\bm{\hat{X}}}}_{2}^{2}+Tr\left(\Sigma_{\bm{X}}+\Sigma_{\bm{\hat{X}}}-2\cdot \sqrt{\Sigma_{\bm{X}}\Sigma_{\bm{\hat{X}}}}\right),
\end{equation}
where $\bm{X}$ and $\bm{\hat{X}}$ are data matrices of real samples and generated samples, respectively, and $\Sigma$ denotes the covariance matrix of a data matrix. In contrast to the Inception score, a lower FID indicates the similarity between the feature distributions since the FID measures a distance.

\section{Methods}
\subsection{Score-guided generative adversarial network (ScoreGAN)}

The main idea of ScoreGAN is straightforward: For its training, the generator in ScoreGAN utilizes an additional loss that can be obtained from the evaluation metric for GANs. Since it has been verified that the evaluation metric strongly reflects the quality and diversity of generated samples \cite{RN1,RN7}, it is expected that the performance of GAN models can be enhanced by optimizing the metrics.

Therefore, the architecture of ScoreGAN corresponds to ControlGAN with an additional evaluator; the evaluator is used to calculate the score, then gradients are backpropagated to train the generator. The other neural network structures are the same as those of ControlGAN.

However, due to the high complexity of GANs, it is not guaranteed that such an approach can work properly, as described in the previous section. Directly optimizing the Inception score can cause overfitting over the network that is used to compute the metric; then, the overfitted GANs produce noises instead of realistic samples even if the score of the generated noise is high \cite{RN12}.

In this paper, we circumvent this problem through two different approaches, i.e., employing the metric as an auxiliary cost instead of the main target of the generator, and adopting another pre-trained network as an evaluator module as a replacement of the Inception network.

\subsubsection{The auxiliary costs using the evaluation metrics}

ScoreGAN mainly uses the ordinary GAN cost in which the adversarial training process is performed while the evaluation metric is utilized as an auxiliary cost. Therefore, the training of the generator in ScoreGAN is conducted by adding the cost of the evaluation metric to (\ref{eq:4}). Such a method using an auxiliary cost has been introduced in ACGAN \cite{RN17}; then, the method has been widely studied in many recent works \cite{RN16}, including ControlGAN \cite{RN18}. As a result of the recent works, it has been demonstrated that the auxiliary costs perform as a ‘rough guide’ for a generator to be trained with additional information. The proposed technique using the evaluation metrics in this paper corresponds to a variant of such a method, where the metrics are used as rough guides to generate high quality and a rich variety of samples. In short, the generator in ScoreGAN aims at maximizing a score in addition to the original cost, which can be represented as follows:
\begin{equation}
\label{eq:8}
    \hat{\bm{\theta}}_{G} = \argmin_{\bm{\theta}_{G}}\left\{ \mathfrak{L}_{G}-\delta\cdot IS\left(\bm{\hat{X}}\right) \right\},
\end{equation}
where $\mathfrak{L}_{G}$ denotes the regular cost for a generator, such as the optimization target in (\ref{eq:4}), $\delta$ is a parameter for the score, and $IS$ is the score that can be obtained from the evaluator. Since (\ref{eq:6}) is differentiable with respect to $G$, $\bm{\theta}_{G}$ can be optimized by the gradients in such a manner.

\subsubsection{The evaluator module with MobileNet}

To obtain the $IS$ in (\ref{eq:8}), originally, the Inception network \cite{RN9} is required as the evaluator in ScoreGAN since the metrics are calculated through the network. However, as described in the previous sections, directly optimizing the score leads to overfitting the network, thereby, making the generator produce noises instead of fine samples. Also, if the Inception network is used for the training, it is challenging to validate whether the generator actually learns features rather than memorizes the network, since the generator trained by the Inception network certainly achieves a high Inception score, regardless of the actual learning.

Therefore, ScoreGAN introduces another network, called MobileNet \cite{RN13}, as the evaluator module, in order to maximize the score. MobileNet \cite{RN13,RN20,RN21} is a comparatively small classifier for mobile devices, which is trained with the ImageNet dataset as well. Due to its compact network size, enabling GANs to be trained with, MobileNet is used in this study. The score is calculated over the feature distribution of MobileNet; then, the generator aims to maximize the score, as described in (\ref{eq:8}). For MobileNet, the pre-trained model in the Keras library is used in this study.

Furthermore, to prevent overfitting MobileNet, ScoreGAN uses a regularized score, which can be represented as follows:
\begin{equation}
\label{eq:9}
    RIS_{mobile}\left(\bm{\hat{X}}\right) := \text{min}\left\{IS_{mobile}\left(\bm{X}\right), IS_{mobile}\left(\bm{\hat{X}}\right)\right\},
\end{equation}
where $RIS$ represents the regularized score, $IS_{mobile}$ denotes the score calculated by the same manner as (\ref{eq:6}) through MobileNet instead of the Inception network. Since a perfect GAN model can achieve a high score that is similar to the score of real data; thus, it is expected that the maximum value of the score that a GAN model can attain is the score of real data. Therefore, such an approach in (\ref{eq:9}) assists the GAN training by reducing the overfitting of the target network.

The evaluation, however, is performed with the Inception network as well as the Inception score, instead of MobileNet and $IS_{mobile}$, which can generalize the performance of ScoreGAN. If ScoreGAN is trained to optimize MobileNet, the training ensures maximizing the score obtained with MobileNet, irrespective of the learning of actual features. Therefore, to validate the performance, the model must be evaluated with the original metric, the Inception score.

Furthermore, the model is further evaluated and cross-validated through the FID. Since the score and the FID measure different aspects of generated samples, the maximization of the score does not guarantee to obtain a low FID. Instead, only if ScoreGAN produces realistic samples that are highly similar to real data in terms of feature distributions, the model can achieve a lower FID than the baseline. Therefore, by using the FID, we can properly cross-validate the model even if the score is used for the target.

\subsection{Network structures and regularization}

\begin{figure*}[t]
\begin{center}
\includegraphics[width=0.9\textwidth]{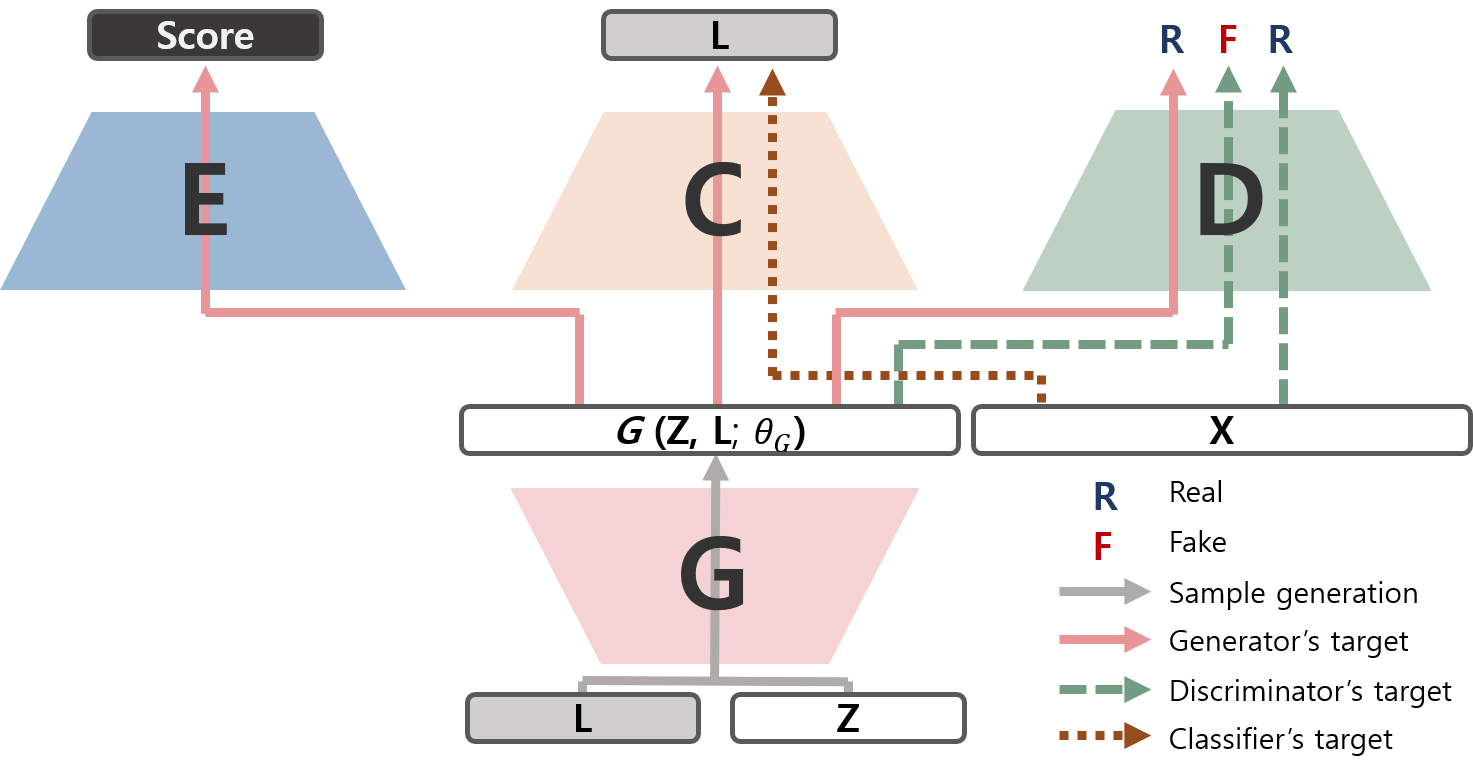}
\caption{\textbf{The structure of ScoreGAN.} The training of each module is represented with arrows. E: Evaluator; C: Classifier; D: Discriminator; G: Generator.}
\label{fig:1}
\end{center}
\end{figure*}

Since ScoreGAN employs ControlGAN structure as the baseline and integrates an evaluator measuring the score with the baseline, ScoreGAN consists of four ANN modules, namely the generator, discriminator, classifier, and evaluator. In short, ScoreGAN additionally uses the evaluator, attached to the original ControlGAN framework. The structure of ScoreGAN is illustrated in Figure \ref{fig:1}.

As described in Figure \ref{fig:1} and (\ref{eq:8}), the generator is trained by targeting the three other ANN modules to maximize the score and minimize the losses, simultaneously. Meanwhile, the discriminator tries to distinguish between real samples and generated samples. The classifier is trained only with real samples in which the data augmentation is applied; then, the loss for the generator can be obtained with the trained classifier. The evaluator is a pre-trained network and fixed during the training of the generator; thereby, the generator learns general features of various objects from the pre-trained evaluator by maximizing the score of the evaluator.

Due to the vulnerable nature of the training of GANs, regularization methods for the ANN modules in GANs are essential \cite{RN22,RN33}. Accordingly, ScoreGAN also uses the regularization methods that are widely employed in various GAN models for its training. Spectral normalization \cite{RN23} and the hinge loss \cite{RN24} that are commonly used in state-of-the-art GAN models are employed in ScoreGAN as well. The gradient penalty with a weight parameter of 10 is used \cite{RN22}. Also, according to recent studies that show the regularized discriminator requires intense training \cite{RN1,RN23}, multiple training iterations for the discriminator is applied; the discriminator is trained over five times per one training iteration of the generator. For the generator and the classifier, the conditional Batch Normalization (cBN) \cite{RN25} and Layer Normalization (LN) \cite{RN26} techniques are used, respectively.

\begin{table*}[tp]
   \small 
\begin{subtable}{0.32\linewidth}
   \begin{tabularx}{\linewidth}{>{\centering\arraybackslash}X }
   \toprule[\heavyrulewidth]\toprule[\heavyrulewidth]
   \textbf{Generator}  \\ 
   \toprule[\heavyrulewidth]
$Z\in \mathbb{R}^{128}$\\
\midrule
Dense $\left(4\times 4\times 256\right)$ \\
\midrule
ResBlock Upsample $\left(256\right)$ \\
\midrule
ResBlock Upsample $\left(256\right)$ \\
\midrule
ResBlock Upsample $\left(256\right)$ \\
\midrule
cBN; ReLU; Conv $\left(3\right)$; Tanh \\
   \bottomrule[\heavyrulewidth] \bottomrule[\heavyrulewidth] 
   \end{tabularx}
\end{subtable}
\begin{subtable}{0.32\linewidth}
   \begin{tabularx}{\linewidth}{>{\centering\arraybackslash}X }
   \toprule[\heavyrulewidth]\toprule[\heavyrulewidth]
   \textbf{Discriminator}  \\ 
   \toprule[\heavyrulewidth]
$Z\in \mathbb{R}^{32\times 32 \times 3}$\\
\midrule
ResBlock Downsample $\left(256\right)$ \\
\midrule
ResBlock Downsample $\left(256\right)$ \\
\midrule
ResBlock $\left(256\right)$ \\
\midrule
ResBlock $\left(256\right)$ \\
\midrule
ReLU; Global Pool; Dense $\left(1\right)$ \\
   \bottomrule[\heavyrulewidth] \bottomrule[\heavyrulewidth] 
   \end{tabularx}
\end{subtable}
\begin{subtable}{0.32\linewidth}
   \begin{tabularx}{\linewidth}{>{\centering\arraybackslash}X }
   \toprule[\heavyrulewidth]\toprule[\heavyrulewidth]
   \textbf{Classifier}  \\ 
   \toprule[\heavyrulewidth]
$Z\in \mathbb{R}^{32\times 32 \times 3}$\\
\midrule
ResBlock $\left(32\right)$ $\times 3$ \\
ResBlock Downsample $\left(32\right)$ \\
\midrule
ResBlock $\left(64\right)$ $\times 3$ \\
ResBlock Downsample $\left(64\right)$ \\
\midrule
ResBlock $\left(128\right)$ $\times 3$ \\
ResBlock Downsample $\left(128\right)$ \\
\midrule
ResBlock $\left(128\right)$ $\times 3$ \\
\midrule
LN; ReLU; Global Pool; \\
Dense $\left(10\right)$ \\
   \bottomrule[\heavyrulewidth] \bottomrule[\heavyrulewidth] 
   \end{tabularx}
\end{subtable}
\centering % center the table
   \caption{\textbf{Architecture of neural network modules.} The values in the brackets indicate the number of convolutional filters or nodes of the layers. Each ResBlock is composed of two convolutional layers with pre-activation functions.} 
\label{tab:1}
\end{table*}

For the neural network structures in ScoreGAN, we follow a typical architecture that is generally introduced in many other studies \cite{RN16,RN27}. The detailed structures are shown in Table \ref{tab:1}. Two Time-scale Update Rule (TTUR) \cite{RN8} is employed with learning rates of $4\times10^{-4}$ and $2\times10^{-4}$ for the discriminator and the generator, respectively. The learning rates halve after 50,000 iterations; then, the models are further trained with the halved learning rates for another 50,000 iterations. The Adam optimization method is used with the parameters of $\beta_{1}=0$ and $\beta_{2}=0.9$, which is the same setting as the other recent studies \cite{RN18,RN23}. The maximum threshold for the training from the classifier is set to 0.1. The parameter $\delta$ in (\ref{eq:8}) that modulates the training from the evaluator is set to 0.5.

\section{Results}
\subsection{Image generation with CIFAR-10 dataset}

The proposed ScoreGAN is evaluated over the CIFAR-10 dataset, which is conventionally employed for a standard dataset to assess the image generation performance of GAN models in many studies \cite{RN15,RN16,RN18,RN23,RN27,RN28,RN29,RN30}. The training set of the CIFAR-10 dataset is composed of 50,000 images that are from 10 different classes. To train the models, we use a minibatch size of 64, and the generator is trained over 100,000 iterations. The other settings and the structure of ScoreGAN that is used to train the CIFAR-10 dataset are described in the previous section. Since the proposed ScoreGAN introduces an additional evaluator compared to ControlGAN, we use ControlGAN as the baseline; thereby, we can properly assess the effect of the additional evaluator.

To evaluate the image generation performance of the models, the Inception score and FID is employed. As described in the previous sections, since the Inception score is an average the relative entropy between each prediction and the marginal predictions, a higher Inception score signifies better quality and rich diversity of generated samples, while a lower FID indicates that feature distributions of generated samples are similar to those of real samples. Notice that, for ScoreGAN, the Inception score and FID are measured after the training iterations (100,000) are completed while we can obtain better score and distance if they are repeatably measured during the training, and then we select the best model among the iterations, as conducted in several studies \cite{RN1,RN27}.

\begin{figure}[t]
\begin{floatrow}
\ffigbox{%
  \includegraphics[width=\linewidth]{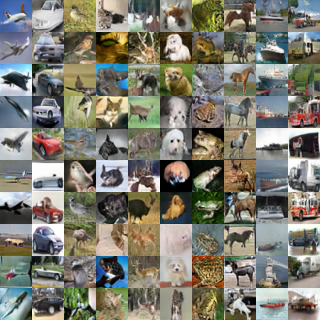}
}{%
  \caption{\textbf{Random examples of generated images by ScoreGAN.} Each column represents each class in the CIFAR-10 dataset.}
  \label{fig:2}
}
\capbtabbox{%
\begin{tabular}{lll}
\toprule[\heavyrulewidth] \toprule[\heavyrulewidth]
Methods & IS & FID \\   [2pt]
\toprule[\heavyrulewidth]
Real data & 11.23$\pm$.20 & - \\   [2pt]
\midrule
ControlGAN \cite{RN18} & 8.61$\pm$.10 & - \\ [2pt]
\makecell[l]{ControlGAN  \\ (w/ Table \ref{tab:1}; baseline)}& 8.60$\pm$.09 & 10.97 \\  [2pt]
\midrule
\makecell[l]{Conditional \\DCGAN \cite{RN28}}  & 6.58 & - \\ [2pt]
AC-WGAN-GP \cite{RN22} & 8.42$\pm$.10 & - \\ [2pt]
CAGAN \cite{RN16} & 8.61$\pm$.12 & - \\ [2pt]
Splitting GAN \cite{RN29} & 8.87$\pm$.09 & - \\ [2pt]
BigGAN \cite{RN1} & 9.22 & 14.73 \\ [2pt]
MHingeGAN \cite{RN27} & 9.58$\pm$.09 & \textbf{7.50} \\ [2pt]
\textbf{ScoreGAN} & \textbf{10.36$\pm$.15} & 8.66 \\ 
\bottomrule[\heavyrulewidth] \bottomrule[\heavyrulewidth] 
   \end{tabular}
}{%
  \caption{\textbf{Performance of GAN models over the CIFAR-10 dataset.} IS indicates the Inception score; FID indicates the Fréchet Inception Distance. The best performances are highlighted in bold.}%
  \label{tab:2}
  }
\end{floatrow}
\end{figure}

Table \ref{tab:2} shows the performance of GAN models in terms of the Inception score and FID. While the neural network architectures of GAN are the same with ControlGAN, the proposed ScoreGAN demonstrates superior performance compared to ControlGAN, which verifies the effectiveness of the additional evaluator in ScoreGAN. The Inception score has increased by 20.5\%, from 8.60 to 10.36, which corresponds to state-of-the-art performance among the existing models thus far. The FID has also decreased by 21.1\% in ScoreGAN compared to ControlGAN in which the FID values of ScoreGAN and ControlGAN are 8.66 and 10.97, respectively. Random examples that are generated by ScoreGAN are shown in Figure \ref{fig:2}.

Such a result validates that the additional evaluator and the auxiliary score in ScoreGAN are effective. It can be said that the generator in ScoreGAN properly learns general features through the pre-trained evaluator and is enforced to produce a variety of samples by maximizing the score, since not only the Inception score has increased, but also the FID has decreased, while the FID measures the similarity between feature distributions, and thus, is less related to the target of ScoreGAN. Also, since ScoreGAN does not use the Inception network as the evaluator and the score, it is hard to regard the generated samples by ScoreGAN as adversarial examples of the Inception network, as shown in the examples in Figure 2 of which samples are far from noises.

\begin{figure*}[t]
\begin{center}
\includegraphics[width=0.9\textwidth]{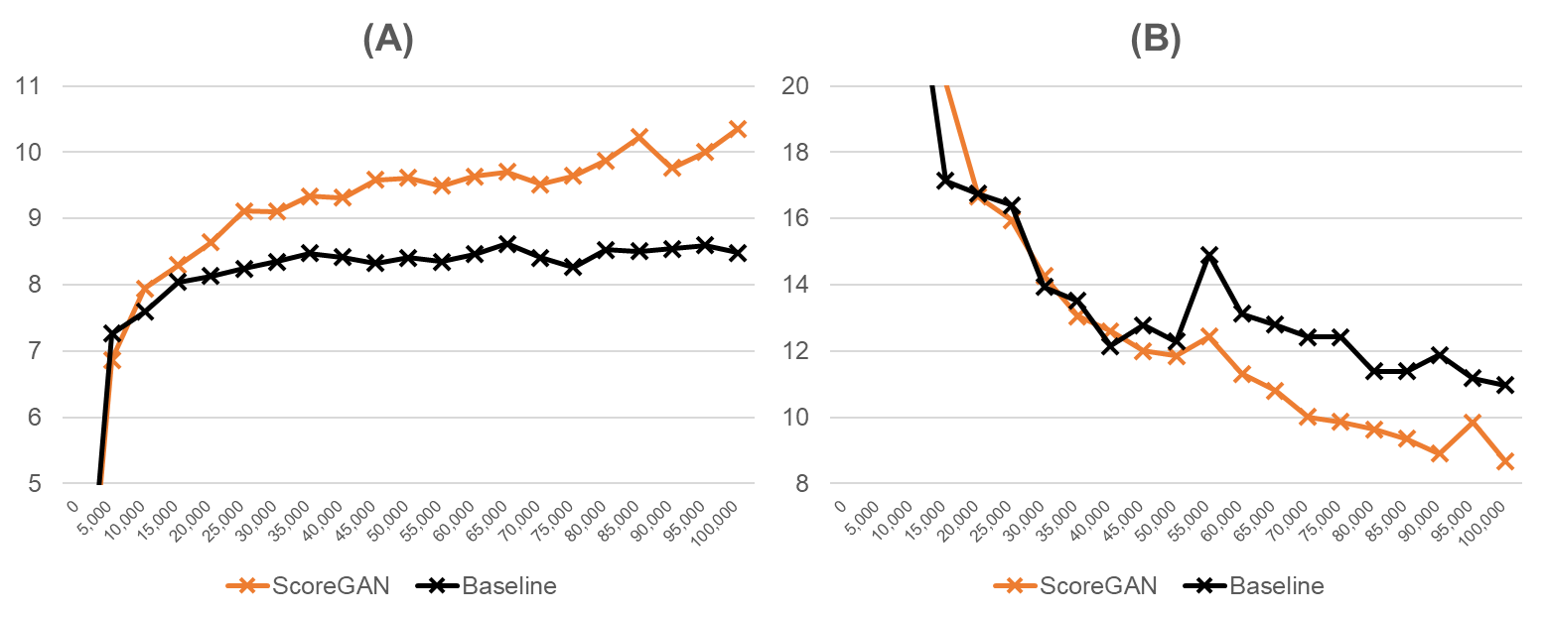}
\caption{\textbf{The performance of ScoreGAN in terms of the Inception score and Fréchet Inception Distance over iterations. (A)} The Inception scores; \textbf{(B)} The Fréchet Inception Distance (FID). The baseline is ControlGAN with the same neural network architecture, identical to that of ScoreGAN.}
\label{fig:3}
\end{center}
\end{figure*}

The detailed Inception score and FID over iterations are shown in Figure \ref{fig:3}. As shown in the figures, the training of ControlGAN becomes slow after 30,000 iterations while the proposed ScoreGAN continues its training. For example, the Inception score of ControlGAN at 35,000 iterations is 8.48, which is 98.6\% of the final Inception score, while, at the same time, the Inception score of ScoreGAN is 9.34, which corresponds to 90.2\% of its final score. The FID demonstrates similar results to those of the Inception score. In ControlGAN, the FID decreases by 10.7\% from 50,000 to 100,000 iterations; in contrast, it declines by 26.9\% in ScoreGAN. Such a result implies that the generator in ScoreGAN can be further trained by the proposed evaluator, although the training of the discriminator is saturated.

\subsection{Image generation with CIFAR-100 dataset}

\begin{table}[t]
\begin{tabular}{lll}
\toprule[\heavyrulewidth] \toprule[\heavyrulewidth]
Methods & IS & FID \\   [2pt]
\toprule[\heavyrulewidth]
Real data & 14.79$\pm$.18 & - \\   [2pt]
\midrule
\makecell[l]{ControlGAN \\ (baseline)} & 9.32$\pm$.11 & 18.42 \\ [2pt]
\midrule
MSGAN \cite{RN31} & - & 19.74 \\ [2pt]
SNGAN \cite{RN30} & 9.30$\pm$.08 & 15.6 \\ [2pt]
MHingeGAN \cite{RN27} & \textbf{14.36$\pm$.09} & 17.30 \\ [2pt]
\textbf{ScoreGAN} & 13.11$\pm$.16 & \textbf{13.98} \\ 
\bottomrule[\heavyrulewidth] \bottomrule[\heavyrulewidth] 
   \end{tabular}
     \caption{\textbf{Performance of GAN models over the CIFAR-100 dataset.} IS indicates the Inception score; FID indicates the Fréchet Inception Distance. The best performances are highlighted in bold.}
     \label{tab:3}
\end{table}

To generalize the effectiveness of ScoreGAN, the CIFAR-100 dataset is employed for the evaluation of GAN models. The CIFAR-100 dataset is similar to the CIFAR-10 dataset, where each dataset contains 50,000 images of size $32\times 32$ in the training set. The difference between the CIFAR-100 dataset and the CIFAR-10 dataset is that the CIFAR-100 dataset is composed of 100 different classes. Therefore, it is generally regarded that the training of the CIFAR-100 dataset is more challenging than that of the CIFAR-10 dataset.

Since existing methods in several recent studies have been evaluated over the CIFAR-100 dataset \cite{RN31}, we compare the performance between ScoreGAN and the existing methods. The performance in terms of the Inception score and FID is demonstrated in Table \ref{tab:3}. The results show that ScoreGAN outperforms the other existing models. While the same neural network architectures are used in both methods, the performance of ScoreGAN is significantly superior to that of the baseline. For instance, the FID significantly declines from 18.42 to 13.98, which corresponds to a state-of-the-art result.

While the Inception score of ScoreGAN is slightly lower than that of MHingeGAN \cite{RN27}, such a disparity results from a difference in the assessment of the scores, in which, for MHingeGAN, the Inception score is continuously measured during the training iterations; then, the best score is selected among the training iterations. In contrast, the Inception score of ScoreGAN is computed only once after 100,000 iterations. Also, in terms of the FID, ScoreGAN demonstrates superior results, compared to MHingeGAN. Furthermore, it is reported that the training of MHingeGAN over the CIFAR-100 dataset collapses before 100,000 iterations.

\section{Conclusion}

In this paper, the proposed ScoreGAN introduces an evaluator module that can be integrated with conventional GAN models. While it is known that the regular use of the Inception score to train a generator corresponds to making noise-like adversarial examples of the Inception network, we circumvent this problem by using the score as an auxiliary target and employing MobileNet instead of the Inception network. The proposed ScoreGAN was evaluated over the CIFAR-10 dataset and CIFAR-100 dataset. As a result, ScoreGAN demonstrated an Inception score of 10.36, which is the best score among the existing models. Also, evaluated over the CIFAR-100 dataset in terms of FID, ScoreGAN outperformed the other models, where the FID was 13.98.

Although the proposed evaluator is integrated with ControlGAN architecture and demonstrated fine performance, it needs to be further investigated whether the evaluator module properly performs when it is additionally used for other GAN models. Since the evaluator module can be employed along with various GANs, the performance can be enhanced by adopting other GAN models. Furthermore, in this paper, only the Inception score is introduced to train the generator while the other metric to assess GANs, i.e., FID, can be used as a score. Such a possibility to use the FID as a score should be further studied as well for future work.

\medskip

\section*{Appendix}
\subsection*{Neural network architectures of ScoreGAN for the CIFAR-100 dataset}

\begin{table*}[h]
   \small 
\begin{subtable}{0.32\linewidth}
   \begin{tabularx}{\linewidth}{>{\centering\arraybackslash}X }
   \toprule[\heavyrulewidth]\toprule[\heavyrulewidth]
   \textbf{Generator}  \\ 
   \toprule[\heavyrulewidth]
$Z\in \mathbb{R}^{128}$\\
\midrule
Dense $\left(4\times 4\times 256\right)$ \\
\midrule
ResBlock Upsample $\left(256\right)$ \\
\midrule
ResBlock Upsample $\left(256\right)$ \\
\midrule
ResBlock Upsample $\left(256\right)$ \\
\midrule
cBN; ReLU; Conv $\left(3\right)$; Tanh \\
   \bottomrule[\heavyrulewidth] \bottomrule[\heavyrulewidth] 
   \end{tabularx}
\end{subtable}
\begin{subtable}{0.32\linewidth}
   \begin{tabularx}{\linewidth}{>{\centering\arraybackslash}X }
   \toprule[\heavyrulewidth]\toprule[\heavyrulewidth]
   \textbf{Discriminator}  \\ 
   \toprule[\heavyrulewidth]
$Z\in \mathbb{R}^{32\times 32 \times 3}$\\
\midrule
ResBlock Downsample $\left(256\right)$ \\
\midrule
ResBlock Downsample $\left(256\right)$ \\
\midrule
ResBlock $\left(256\right)$ \\
\midrule
ResBlock $\left(256\right)$ \\
\midrule
ReLU; Global Pool; Dense $\left(1\right)$ \\
   \bottomrule[\heavyrulewidth] \bottomrule[\heavyrulewidth] 
   \end{tabularx}
\end{subtable}
\begin{subtable}{0.32\linewidth}
   \begin{tabularx}{\linewidth}{>{\centering\arraybackslash}X }
   \toprule[\heavyrulewidth]\toprule[\heavyrulewidth]
   \textbf{Classifier}  \\ 
   \toprule[\heavyrulewidth]
$Z\in \mathbb{R}^{32\times 32 \times 3}$\\
\midrule
ResBlock $\left(32\right)$ $\times 3$ \\
ResBlock Downsample $\left(32\right)$ \\
\midrule
ResBlock $\left(64\right)$ $\times 3$ \\
ResBlock Downsample $\left(64\right)$ \\
\midrule
ResBlock $\left(128\right)$ $\times 3$ \\
ResBlock Downsample $\left(128\right)$ \\
\midrule
ResBlock $\left(256\right)$ $\times 3$ \\
\midrule
LN; ReLU; Global Pool; \\
Dense $\left(100\right)$ \\
   \bottomrule[\heavyrulewidth] \bottomrule[\heavyrulewidth] 
   \end{tabularx}
\end{subtable}
\centering % center the table
   \caption{\textbf{Architecture of neural network modules for the training of the CIFAR-100 dataset.} The values in the brackets indicate the number of convolutional filters or nodes of the layers. Each ResBlock is composed of two convolutional layers. The difference between the architecture for the CIFAR-10 dataset is at the classifier, in which 256 filters are used in the last three ResBlocks.} 
\label{tab:4}
\end{table*}

\newpage 
\subsection*{Generated samples by ScoreGAN trained with the CIFAR-100 dataset}

\begin{sidewaysfigure}[H]
    \includegraphics[width=0.98\textwidth]{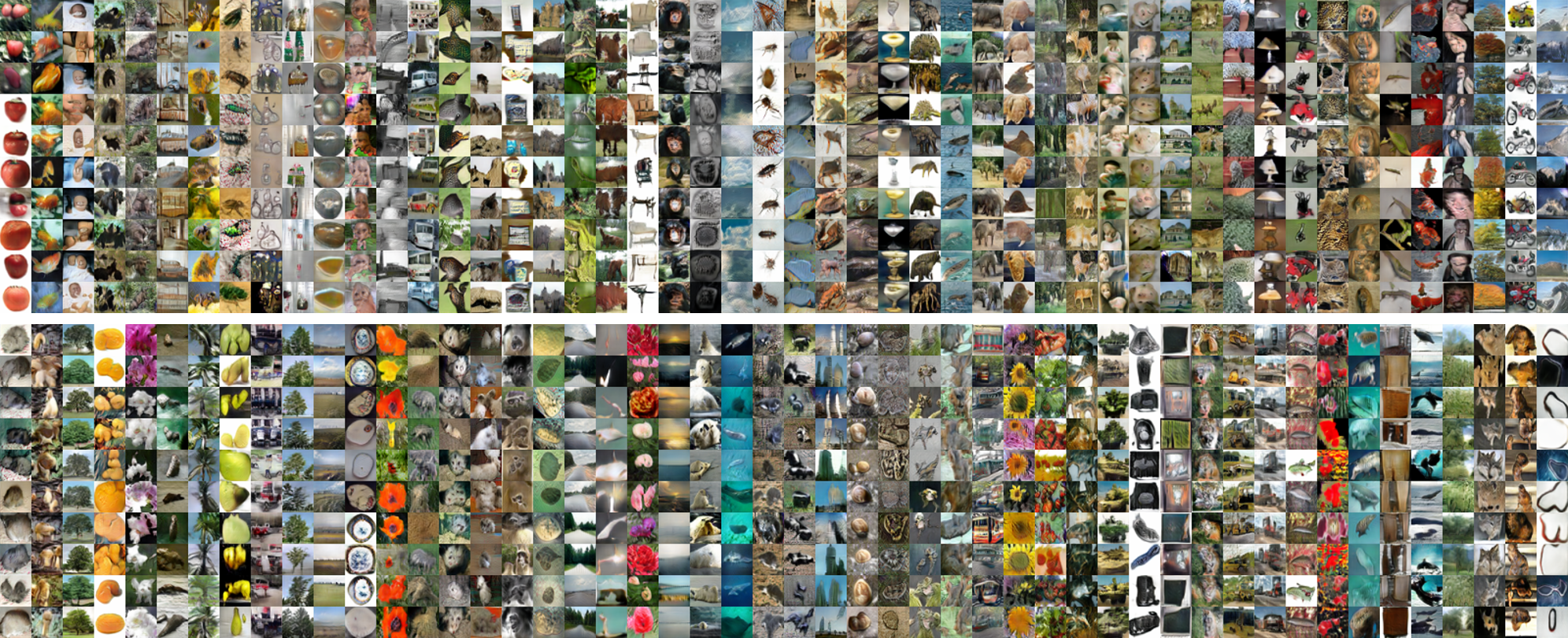}
    \caption{\textbf{Random examples of generated images by ScoreGAN.} Each column represents each class in the CIFAR-100 dataset.}
    \label{fig:4}
\end{sidewaysfigure}

\newpage 
\small

\bibliography{refs}
\bibliographystyle{IEEEtran}

\end{document}